\documentclass[a4paper,twoside]{article}

\usepackage{graphicx}
\usepackage{epsfig}
\usepackage{subcaption}
\usepackage{calc}
\usepackage{amssymb}
\usepackage{amstext}
\usepackage{amsmath}
\usepackage{amsthm}
\usepackage{multicol}
\usepackage{pslatex}
\usepackage{apalike}
\usepackage{SCITEPRESS}     

\begin{document}

\title{CHALLENGES IN DESIGNING DATASETS AND VALIDATION FOR AUTONOMOUS DRIVING}

\author{\authorname{Michal U\v{r}i\v{c}\'{a}\v{r}\sup{1}, David Hurych\sup{1}, Pavel K\v{r}\'{i}\v{z}ek\sup{1} and Senthil Yogamani\sup{2}}
\affiliation{\sup{1}Valeo R\&D DVS, Prague, Czech Republic}
\affiliation{\sup{2}Valeo Vision Systems, Tuam, Ireland}
\email{\{michal.uricar, david.hurych, pavel.krizek\}@valeo.com, senthil.yogamani@valeo.com}
}



\abstract{
Autonomous driving is getting a lot of attention in the last decade and will be the hot topic at least until the first successful certification of a car with Level~5 autonomy~\cite{SAE_automation}. There are many public datasets in the academic community. However, they are far away from what a robust industrial production system needs. There is a large gap between academic and industrial setting and a substantial way from a research prototype, built on public datasets, to a deployable solution which is a challenging task. In this paper, we focus on bad practices that often happen in the autonomous driving from an industrial deployment perspective. Data design deserves at least the same amount of attention as the model design. There is very little attention paid to these issues in the scientific community, and we hope this paper encourages better formalization of dataset design. More specifically, we focus on the datasets design and validation scheme for autonomous driving, where we would like to highlight the common problems, wrong assumptions, and steps towards avoiding them, as well as some open problems.
}

\onecolumn \maketitle \normalsize \setcounter{footnote}{0} \vfill

%


\section{INTRODUCTION}\label{sec:intro}

\begin{figure*}[tb]
    \centering
    \includegraphics[width=0.9\linewidth]{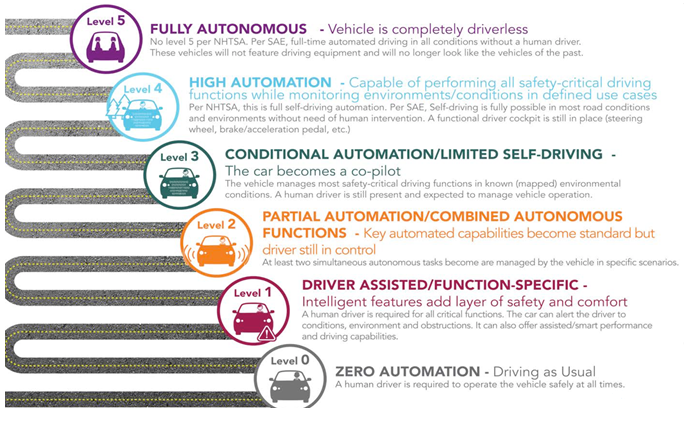}
    \caption{Levels of autonomy as described by~\cite{SAE_automation}. Note, that the standard in $2018$ is level~2, since the first sensor allowing level~3 functionality was just released in $2018$. We can participate quick adaptation to level~3 in the next year. The ultimate goal is, of course, level~5, for which a lot of companies is aiming now. However, the path to get to the level~5 is still quite long and the predictions are speculating about decades to get there.}
    \label{fig:SAE_levels}
\end{figure*}

We have the privilege to live in the exciting era of high pace research and development aiming for the full autonomy in the ground transportation, involving all major automotive industries. Nowadays, the sort of standard is the autonomy level~2~\cite{SAE_automation}. We can see the progress towards levels~3 and 4, and the ultimate goal is, of course, to achieve level~5, i.e., the real full autonomy. In Figure~\ref{fig:SAE_levels}, we outline all the levels of autonomy in automotive for reference, as described by~\cite{SAE_automation}.

Naturally, there is a high motivation and willingness to speed up the progress in combination with the recent success of deep neural networks. However, this leads to the development of certain bad practices, which are progressively more and more visible in research papers. The goal of this paper is to determine some of the bad practices, especially those related to the issue of dataset design and validation scheme, and propose the ideas for fixing them. Apart from that, we would also like to identify several open problems for which the standardized solution is yet to be discovered. 

The importance of dataset design is often overlooked in the computer vision community;
the problem was addressed in detail in the ECCV workshop in 2016~\cite{eccv16datasets}.
In~\cite{khosla2012undoing} the authors discuss issues with dataset bias and how to
address them. In general, we can say that having a good and representative data is the
crucial problem of virtually all machine learning techniques. Often, the applied
algorithms come together with a requirement for data to be independent and identically
distributed  (i.i.d). However, this requirement is frequently broken and not checked
for. Either the dataset parts are obtained from different distributions or their
independence is questionable. Also the definition of terms \textit{identically} and
\textit{independently} depends on the foreseen application.

Frequently, we can also see that researchers blindly follow the results of the evaluation represented by a key performance indicator (KPI) without making the trouble to check for the experiment correctness. With the increasing size of the datasets, the errors in annotations become significant, and the absence of a careful inspection might be dangerous. Especially, when we achieved the state where the improvements in decimals point are considered essential. A lot of models are discarded during the design, because of the lack of systematical analysis and the effort of getting a real insight. However, such a complete and systematical analysis might render some of these models more robust or to better generalize across different datasets. 

\begin{figure}
    \centering
    \includegraphics[width=\linewidth]{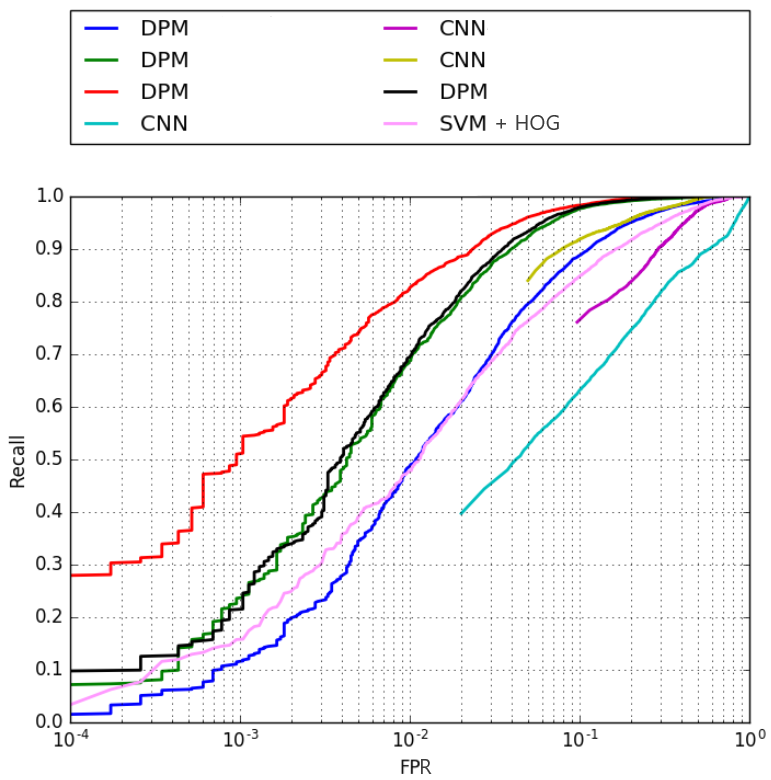}
    \caption{Performance comparison for equally complex CNN and DPM models on a pedestrian detection task. Multiple settings were tested in an attempt to reach best performance for both methods. Due to internal confidentiality rules we can not reveal individual settings. Here only SVM+HOG had lower complexity than other models and served as a baseline. 3 curves do not pass through the whole [Recall, FPR] space because of lack of data-points in next confidence step. We were interested in parts of curves with recall higher than 0.9.}
    \label{fig:dpm}
\end{figure}

Last but not least, we should emphasize the importance of fair comparisons regarding the used resources or a model complexity. Taking deep neural networks as a gentle example, what if we use an ensemble of simpler classifiers, trying to call the complexity of a neural net? Would we still see the same performance gap? As a typical example of methods and models compared only from the performance point of view while ignoring the computational complexity we may take deep Convolutional Neural Networks (CNN)~\cite{KrizhevskySH12} versus the Deformable Part based Models (DPM)~\cite{felzenswalb2010}. 
The CNN models are dominating DPM since their break-through in ImageNet Large Scale Visual Recognition Challenge (ILSVRC)~\cite{ILSVRC15} back in 2012. However, we tried to do a bit ``fairer'' comparison, where the model complexity was more or less matched (i.e. both CNN and DPM models consisted of the similar number of operations in the inference calculation).
The results from the independent testing set are summarized in Figure~\ref{fig:dpm}, where we experimented with these methods on our internal pedestrian dataset in real driving scenarios. 
Both CNN and DPM models were trained on the same training set. The dataset splits were obtained by the means of stratified sampling. The DPM model is better in several different settings. 

The rest of the paper is structured as follows. Section~\ref{sec:dataset_design} discusses the current bad practices in dataset design, emphasizes its importance and lists open issues. Section~\ref{sec:validation} summarizes issues related to the validation of a safety system. Finally, Section~\ref{sec:conclusions} concludes the paper and provides future directions.


\section{DATASET DESIGN} \label{sec:dataset_design}

Typically in academia, test and validation datasets are provided, and the goal is to get the best accuracy. However, in the industry deployment setting, datasets have to be designed interactively along with the model design. Unfortunately, there is minimal systematic design effort as it is difficult to formulate the problem and quantify the quality of datasets. First, there is a data capture process where cameras mounted on a vehicle capture necessary data. This process has to be repeated across many scenarios like different countries, weather conditions, times of the day, etc.

\begin{table*}
    \centering
    \caption{Popular automotive datasets for semantic segmentation.}
    \label{tab:popular_datasets}
    \begin{tabular}{|l|c|c|c|}
        \hline
        \textbf{Dataset} & CamVid & Cityscapes & Synthia \\
        & \cite{brostow2008segmentation} & \cite{cordts2016cityscapes} & \cite{ros2016synthia} \\
        \hline 
        \textbf{Annotation} & $700$ images & $5000$ images & $200,000$ images\\
        \hline
        \textbf{Note} & Cambridge captures & Germany captures & Synthetic data \\
        \hline\hline 
        \textbf{Dataset} & Virtual KITTI & Mapillary Vistas & ApolloScape \\
        & \cite{gaidon2016virtual} & \cite{neuhold2017mapillary} & \cite{huang2018apolloscape} \\
        \hline
        \textbf{Annotation} & $21,260$ & $25,000$ images; $100$ classes & $143,000$ images; $50$ classes \\
        \hline
        \textbf{Note} & Synthetic data & Six continents & China captures \\
        \hline
    \end{tabular}
\end{table*}

Table~\ref{tab:popular_datasets} lists the popular automotive datasets for semantic segmentation task. There is visible progress towards the increased number of samples as well as a change in direction to realistic data. 

\subsection{Typical Scenarios of Dataset Design in Autonomous Driving}

There are three scenarios currently possible for automated driving researchers: (i) academic setup where public datasets are used as they are. However, these have commercial licensing restrictions and cannot be used freely for industrial research nor for final production, (ii) proof of concept setup where dataset is collected for a restricted scope, e.g. one city, regular weather conditions, and (iii) a production system where dataset has to be designed for all scenarios, e.g., a large set of countries, weather conditions, etc. 

Here, we focus our discussion on the third type. A typical process for dataset design comprises of the following steps. Firstly, the requirements are created for coverage of countries, weather conditions, object diversity, etc. Then video captures are acquired and their frames arbitrarily sampled without a systematic sampling strategy. Next step is creating the training, validation and test splits from all gathered images. This can be done either randomly, or in a better case by stratified sampling, retaining the class distributions among the splits. Then, after the model is trained, one has to evaluate the KPIs, such as the mean intersection over union (mIoU). Last but not least step is a na\"ive search for corner cases in the test split and addition of such infected samples to the training split (in best case by obtaining new data, which look similar to the testing ones). 


\section{VALIDATION} \label{sec:validation}

Validation scheme is another critical part of the current research. In the automotive industry, this topic is even more critical, due to the very stringent requirements on safety. However, we would like to emphasize here, that the common problems with validation are shared among other fields as well. 

The AD systems have unique criteria due to functional safety and traceability issues. The artificial intelligence software for the AD has to comply with strict processes like ISO26262 to ensure functional safety. Thus, apart from accuracy validation, it is essential to do rigorous testing of software stability. However, unit testing of AI algorithms comes with additional challenges, like large dimensionality of data, or the abstract nature of the model and automated code generation. Due to these challenges, it is difficult to write tests manually. There have been attempts to generate tests automatically using deep learning like DeepXplore~\cite{pei2017deepxplore} and DeepTest~\cite{tian2018deeptest} focused on the AD.

\subsection{The Need for Virtual Validation and Their Limitations} \label{sec:virtual_validation}

In the automotive industry, one tests the algorithms by recording a lot of hours of various scenarios which should ideally cover all possible real-world situations. Afterward, these hours of recordings are statistically evaluated, and the algorithm is allowed to go for the start of production only if fulfilling the strict requirements, which were formulated and ideally also fixed at the beginning of the project. Despite sounding completely legitimate at first, such an approach has several important flaws. 
The first, and probably also the most important one is the physical and practical impossibility to cover all real-world scenarios. Let us formulate an example explaining our claims: we want to test the automated parking functionality enhanced by pedestrian detection algorithm. The vehicle should send a command for braking if there is a pedestrian within critical distance and on the collision course present. Now, imagine a legitimate scenario, where the pedestrian should be a toddler sitting right behind the car. In most of the countries across the whole globe, their laws prohibit making of such recordings.
The second problem is in setting up the requirements at the beginning of the project. If we agree that it is not possible to record all real-world situations, freezing the requirements tends to influence the scenarios to record as well as their complexity. 

One might say, that the solution is obvious--- use virtual validation or some other workaround, like usage of dummy objects. However, doing so bring other problems, such as the realism of these artificial scenarios. It is clear that we will never be able to evaluate it for some of the real-world situations.


A common problem (not only in the automotive industry) is the design of the validation scheme itself. Typically, we can see that the algorithm was optimized for a specific criterion, using a particular loss function. Ideally, we should see the same loss function used in the evaluation, alas, it is not uncommon to see something not even similar to the original loss used in testing. The problem with this setup is that it is not possible to optimize for a beforehand unknown criterion. 

\subsection{Model Survivorship Bias}

This mistreated optimization is connected to another significant problem--- overfitting to (not only) standard datasets. In research, but also in industry, only the models which are obtaining the best results are reported or survive. Nobody tells, how many times the model failed on those data before it was tweaked enough to provide the best results. Not reporting the negative results is counter-productive~\cite{Borji18}. With the increase of the deep learning models in the game, this problem is even worse. Deep networks are known to easily fit random labels, even for randomly generated images~\cite{ZhangBHRV16}. 

\subsection{Complete Reporting and Replicability}

Another, and unfortunately also very frequent, problem is that certain important statistics are not being reported. Only rarely, we can see that authors of the research paper give away also their splits of datasets to training/validation/testing parts. Quite frequently, they do not even bother mentioning the key statistics, such as the number of samples they used, or how the splits were obtained. This problem is connected with a choice of machine learning method, which is not well justified. For example, a lot of techniques come with an assumption about the data distribution. Important and integral part of each experimental evaluation should be its replicability by a non-involved party. 

\subsection{Cross-validation and the Law of Small Numbers}

The emphasis on early deployment in the automotive industry often leads to unjustified design choices, which do not have support in data. The infamous law of small numbers gets into the practice. Usually, the researchers and developers have to deal with the insufficient amount of data to do a statistical evaluation. Then, due to the lack of time, some early decisions are made. The smaller the sample of data was used for performing the evaluation, the higher the probability of wrong outcomes. Just imagine you have a fair coin, so the expected probability of getting head after flipping it is 0.5. Let us conduct the following experiment. Flip the coin ten times and count the number of heads/tails. Very likely, you will not get the five heads. Now flip the same coin a thousand times, marking the number of heads. This time, the number of heads will be close to five hundred. If one does only the first experiment, he might be in temptation to question the fairness of the coin. While after the second one, such conduct feels unjustified. It is common to see quite a small number of samples for some particular tasks, as well as only one evaluation over a single split of data.


\subsection{The Need for Customized Evaluation Metrics}

\begin{figure*}[tb]
    \centering
    \includegraphics[width=\linewidth]{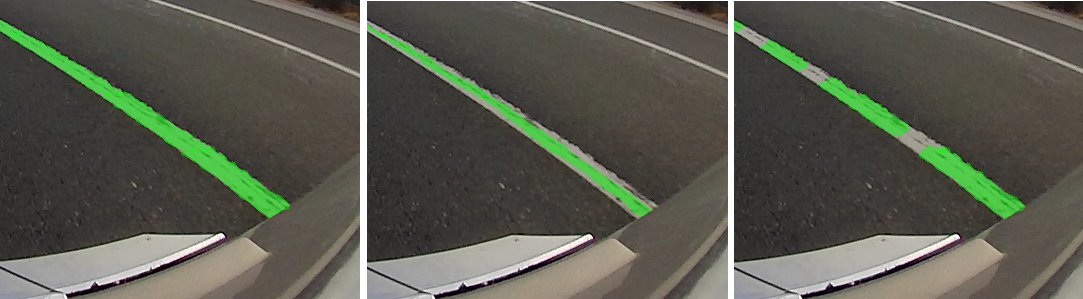}
    \caption{Wrong metrics can lead to misinterpretation of the results. Left: ideal segmentation of the lane marking--- whole line is correctly covered. Middle: one possible case of a realistic segmentation obtained from a learned model. Note that the segmented marking is thinner than it should be. However, it is continuous non-interrupted line as it should be. Right: another possible case of a realistic segmentation. This time, the width of the lane marking is correctly segmented. However, there are interruptions, which render the marking looking like a dashed line. Note, that such misinterpretation might be grievous, since as we know both type of markings are admissible having completely different semantic meaning. The example on the rightmost image has higher IoU value, than the middle one.}
    \label{fig:IoU_critique}
\end{figure*}

Standard performance measures, such as mIoU for semantic segmentation, may not translate well to the end user application needs. Let us take the automated parking functionality with pedestrian detection as an example again. A perfect segmentation of a pedestrian is not necessary, and just a coarse detection is sufficient for initiating the braking. Another example is recognition of the lane markings--- there are nice examples, where a higher mIoU does not necessarily lead to a better segmentation of the main shape of the marking, which is crucial for its recognition. In Figure~\ref{fig:IoU_critique}, we depict one of such examples. In both of these cases, custom tailored evaluation metric is the key for a better algorithm. And visual checks of the results are a must.

\section{DISCUSSION} 
\label{sec:discussion}

In this section, we would like to discuss several important open issues and suggestions for improvement.

\subsection{Open Issues and Suggestions for Improvement}


Many visual perception tasks, like semantic segmentation, need a very expensive annotation,
leading to unnecessarily smaller datasets. Synthetic datasets, like~\cite{ros2016synthia},
\cite{gaidon2016virtual}, \cite{DosovitskiyRCLK17}, \cite{MuellerDGK18},
can be useful as potential mitigation of the lack of data. However, domain adaptation is
usually required, and it is not clear what ratio of synthetic to real data would be still
beneficial. 

Due to the popularity of the AD, the available datasets are snowballing, and the choice of datasets starts to be a problem itself. Moreover, in research, there is no synchronization on datasets, and it is difficult to compare different works justly. It might be helpful if the community agrees on some standardized dataset (combining the strengths and weaknesses of all of them) to have a possibility to compare algorithms more thoroughly and honestly. For example, there is no available dataset with wide-angle fisheye camera images. Such a camera is a standard in the AD for capturing the $360^\circ$ view around the vehicle. A publicly available dataset with multiple cocoon cameras, which are typical for the AD, is also missing. 


An automated sampling mechanism for acquiring the training images the goal of which is to get rid of redundant samples and providing a maximal diversity is an open problem. The dataset and model design is done iteratively, and samples are added on the go to improve the model performance. This process is dangerous since such an approach might easily break the i.i.d. requirements on data. A corner case mining is a related topic, where difficult samples are identified, and their knowledge is used for improving the performance. Note, that this process usually takes into the account the sample similarity in the image space. However, one would benefit if the mining would be based on the similarity measured in the classifier's feature space. 

\begin{figure}[tb]
    \centering
    \includegraphics[width=\linewidth]{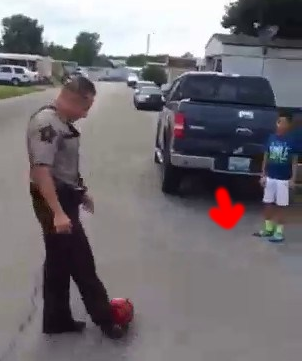}
    \caption{Typical situation, which is required to be covered by data, but which is also prohibited to arrange for recording by law--- a child is playing with a ball, focusing on the play and not paying any attention to the car, which is parking and on the collision path with the child.}
    \label{fig:parking_kid_ball}
\end{figure}

Data augmentation is an important mechanism to obtain samples for difficult or rare scenarios. We can take an automated parking system with pedestrian detection as an example. We would like to have data where children are playing with a ball and sometimes blindly follow the ball which gets on the collision path with the vehicle. One such possible situation is depicted in Figure~\ref{fig:parking_kid_ball}. It is clear that we cannot record such scenario, due to the safety and legal issues. We see the possibility in bypassing such scenario by recording in a controlled environment and applying GANs~\cite{GoodfellowPMXWOCB14} for the domain transfer to fit the AD needs~\cite{chan2018everybody},~\cite{pmlr-v80-hoffman18a}. 

\section{CONCLUSIONS} \label{sec:conclusions}

In this paper, we attempt to emphasize the importance of dataset design and validation for the AD systems. Both dataset design and validation are highly overlooked topics which have created a large gap between academic research and industrial deployment setting. There is a considerable effort to go from a model which achieves state-of-the-art results in an academic context to the development of a safe and robust system deployed in a commercial car. Unfortunately, there is very little scientific effort spent in this direction. We have tried to summarize the bad practices and listed open research problems based on our experience in this area for more than ten years. Hopefully, this encourages further scientific research in this area and places a seed for future improvement.



{\small
\bibliographystyle{apalike}
\bibliography{refs.bib}
}

\end{document}